\documentclass[
]{ceurart}

\begin{document}

\copyrightyear{2022}
\copyrightclause{Copyright for this paper by its authors. Use permitted under Creative Commons License Attribution 4.0 International (CC BY 4.0)}

\conference{The 17th International Workshop on Ontology Matching, The 21st International Semantic Web Conference (ISWC) 2022, 23 October 2022, Hangzhou, China}

\title{GraphMatcher: A Graph Representation Learning Approach for Ontology Matching}

\author{Sefika Efeoglu}[%
orcid=,
email=sefika.efeoglu@fu-berlin.de
]

\address{Free University of Berlin, Department of Computer Science, Takustraße 9, 14195 Berlin, Germany}

\begin{abstract}
Ontology matching is defined as finding a relationship or correspondence between two or more entities in two or more ontologies. To solve the interoperability problem of the domain ontologies, semantically similar entities in these ontologies must be found and aligned before merging them. GraphMatcher, developed in this study, is an ontology matching system using a graph attention approach to compute higher-level representation of a class together with its surrounding terms. The GraphMatcher has obtained remarkable results in in the Ontology Alignment Evaluation Initiative (OAEI) 2022 conference track. Its codes are available at ~\url{https://github.com/sefeoglu/gat_ontology_matching}.
\end{abstract}


\maketitle

\section{Presentation of the system}
GraphMatcher is a new ontology matching system based on graph representation learning using a graph attention~\cite{GAN_2018} together with a new neighbourhood aggregation approach. The graph representation learning approach has leveraged the graph attention and introduces a new neighbourhood aggregation algorithm that increases the contextual information of the centre class and property.

\subsection{Proposal and general statement}
Ontology matching is to find a relationship or correspondence between two or more entities in two or more independent ontologies. The alignments of two ontologies are classified as two different cases: (i) simple alignment and (ii) complex alignment~\cite{Thiblin2020SurveyOC}. The simple alignment is defined as the mapping of the class names according to word-based similarity, while complex alignment considers the meaning of two classes to decide whether they are similar~\cite{Thiblin2020SurveyOC}. To understand the meaning of a class (sequence), the contextual information of the class or property is needed. In this case, we must decide which neighbour contributes to their contextual information.

Many logic- and algorithm-based ontology matching tools, such as LogMAP~\cite{LogMap} and AML~\cite{AML}, solve this interoperability problem of domain ontologies by using algorithms and logic-based approaches. In addition to these approaches, DeepAlignment~\cite{kolyvakis-etal-2018-deepalignment}, VeeAlign~\cite{VeeAlign}, and Convolutional Networks of Bento et al. (2020)~\cite{bento-etal-2020-ontology} apply machine learning (ML) for matching. Nevertheless, according to OAEI's~\footnote{Ontology Alignment Evaluation Initiative (OAEI):\\ \url{http://oaei.ontologymatching.org/}} conference track results, these ML approaches cannot achieve a better performance than traditional tools such as AML and LogMAP. The weakness of these ML approaches might be due to the lack of contextual information about the property and class. Another limitation is in how to represent the ontology's data as a convolutional graph - such as an image in which each pixel in the image data has the same number as its neighbouring pixels - whereas each class in ontology has a different number of its neighboring terms like an arbitrary graph~\cite{GAN_2018}. The most appropriate way of representing the data in the ontology is the arbitrary graph.

Since the ontology represents the data with the arbitrary graph, we aim to develop a graph representation learning model based on a graph attention mechanism~\cite{GAN_2018} using Siamese networks~\cite{bento-etal-2020-ontology,adol1, VeeAlign} to find the semantically similar concepts within the ontologies. The graph attention mechanism computes the higher-level representation of a concept and its surrounding concepts (features). The model then finds similarity scores between the concept pairs among the aligned ontology pairs and determines the concept alignments.

\subsection{Specific techniques used}

GraphMatcher utilises a graph representation learning approach that uses the graph attention~\cite{GAN_2018}. Its network consists of five layers. The main contribution is the adaptation of the graph attention to the Siamese network in the third layer.
\begin{figure}[!h]
\centering
\includegraphics[width=12cm, height=5cm]{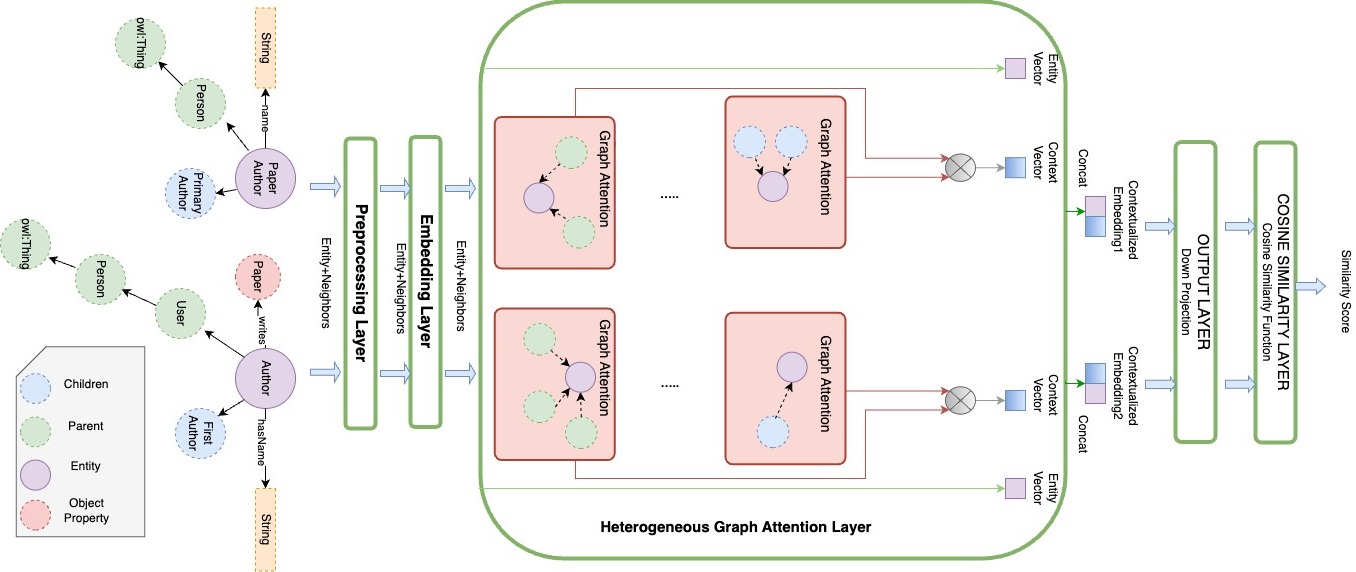}
\caption[Graph Neural Network Architecture]{The proposed network~\footnotemark is an application of heterogeneous graph attention on Siamese networks to find the similar classes.}
\label{fig:siamhan}
\end{figure}
\footnotetext{The orders of the layers in the network is similar to the VeeAlign~\cite{VeeAlign}, since this work's extended version has also increased the performance of the VeeAlign with our neighborhood aggregation algorithm in `` Efeoglu, S. (2021). A Deep Learning Approach for Domain-Specific Ontology Construction. University of Potsdam. [Master's Thesis]''.}
\subsubsection{Preprocessing} Data preprocessing is one of the most significant parts of developing a ML model and is required to explain the variability of features in a sample. In this study, we have handled data preprocessing of an ontology in six steps: (i) an ontology parsing, (ii) tokenization, (iii) finding the abbreviations, (iv) cleaning from stop words, (v) neighbourhood aggregation for creating the context, and (vi) finding the embedding of the terms.

\subsubsection{Embedding layer} 
We used the pre-trained Universal Sentence Encoder (USE) to obtain the word embedding vector of the class and property. According to benchmark results~\cite{Hassan2019BERTEU}, the USE outperforms the BERT encoder in semantic sentence encoding. Therefore, we chose the USE for word embeddings.

\subsubsection{Heterogeneous graph attention layer}
The graph attention introduced in graph attention networks is used to cluster and classify the citation graphs~\cite{GAN_2018}. It provides inductive and transactive learning approaches. Its inductive learning approach computes contextual information by applying an attention centred on the centre class in the neighbourhood graph in our system. The centre class has five homogeneous graphs consisting of its neighbouring terms, and these five graphs refer to one heterogeneous graph.
\begin{figure}[!h]
\begin{center}
    \includegraphics[height=6cm, width=10cm]{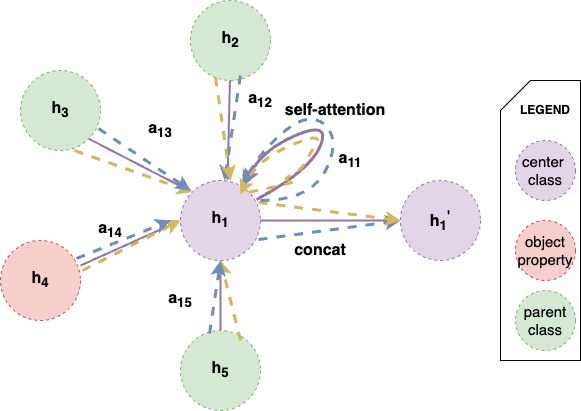}
\end{center}

    \caption[Representation of graph attention on a homogeneous graph: `subClassOf' relationship]{This figure shows how graph attention is applied to a homogeneous graph. The relationship between the centre class and its neighbouring classes is the same in the homogeneous graphs such as `subClassOf'. In this graph, the `ObjectProperty' node represents the restrictions belonging to the definition of a child or parent class. Bag-Of-Word is used to represent the centre class and its neighbouring classes. $h_{1}^{'}$ shows the contextual information of $h_{1}$ in terms of this `subClassOf' relation after applying graph attention. In addition, a$_{ij}$ denotes $\alpha_{ij}$ in Equation~\ref{eq:gat_5}.}
 \label{fig:gat_graph}
\end{figure}

The relation between the centre class and other terms in each homogeneous graph is the same. For example, the relation between the centre node and other terms in Figure~\ref{fig:gat_graph} is `subClassOf'. However, some homogeneous graphs having `equivalentClass' or `subClassOf' relationships might contain some terms regarding restrictions defined with `ObjectProperty' in the neighbourhood aggregation. The figure explains how the centre class' contextual information is computed by the graph attention in one of its homogeneous neighbourhood graphs. The main difference between the original graph attention approach and our graph attention is that this attention mechanism is applied to the heterogeneous graph containing five homogeneous subgraphs.
The attention in this layer runs the following various attention mechanisms~\cite{GAN_2018}.

A set of features (the centre class and its neighbouring terms), which is  inputs of the graph attention layer, is denoted in Eq.~\ref{eq:gat_1} where $\vec{h_{i}} \epsilon \mathbb{R}^{F}$.

\begin{equation}
h= \left \{ \vec{h_{1}}, \vec{h_{2}}, \vec{h_{3}}......\vec{h_{N}}, \right \}
\label{eq:gat_1}
\end{equation}

The layer converts the input features to the new higher-level representation of the feature list like defined in Eq.~\ref{eq:gat_2} where $\vec{h_{i}^{'}} \epsilon \mathbb{R}^{F^{'}}$
\begin{equation}
    h^{'}= \left \{ \vec{h_{1}^{'}}, \vec{h_{2}^{'}}, \vec{h_{3}^{'}}......\vec{h_{N}^{'}}, \right \}
    \label{eq:gat_2}
\end{equation}
Using features in Eq.~\ref{eq:gat_1}, we have obtained the higher-level representation of the class' neighbours, namely its contextual information in this layer.
The new features are computed by Eq.~\ref{eq:gat_3}, and
$\sigma$ indicates the linear activation function like sigmoid or softmax in the Eq.~\ref{eq:gat_3}.
To indicate the higher-level representation of the set of features, $W \epsilon \mathbb{R}^{F' X F}$ is used as a learnable parameter, and shared linear transformation is applied to each feature. ``K'' is the number of independent heads, and ``K'' equals five in our system.
\begin{equation}
h' = ||_{k=1}^{K} \sigma \left (\sum\alpha^{k}_{ij}W^{k}\vec{h_{j}}  \right )
\label{eq:gat_3}
\end{equation}
$\alpha$ $\epsilon$ $\mathbb{R}^{F}_{'}$ X $\mathbb{R}^{F}_{'}$ 
denotes a shared attention mechanism and a layer using the self-attention. The following equation computes attention coefficients ($e_{ij}$):
\begin{equation}
e_{ij} = \alpha(W\vec{h}_{i}, W\vec{h}_{j}) 
\label{eq:gat_4}
\end{equation}

To compute the $\alpha_{ij}$ for $h'$ in the Eq.~\ref{eq:gat_2}, this coefficient attention mechanism is applied in Eq.~\ref{eq:gat_5} where $\vec{a} \epsilon \mathbb{R}^{2F}$, and $\alpha_{ij}$ is attention mechanism parameterized by $\vec{a}$ weight vector.
\begin{equation}
    \alpha_{ij} = softmax(e_{ij}) = \frac{\exp (LeakyReLU(\vec{a}{_{}}^{T} [W\vec{h_{i}}||W\vec{h_{j} }]))}{\sum_{k \epsilon N_{i}}^{}\exp (LeakyReLU(\vec{a}{_{}}^{T} [W\vec{h_{i}}||W\vec{h_{k} }])) } 
    \label{eq:gat_5}
\end{equation}

Using the formula in the Eq.~\ref{eq:gat_3} in this layer, we have obtained the higher-level representation of the class' neighbours, namely its contextual information in this layer.

\subsubsection{Output layer}
The output layer provides down sampling (dimensional reduction) of the contextual information, which is the concatenation of the class embedding and higher representations of the class' neighbours.

\subsubsection{Cosine similarity layer}
The cosine similarity layer measures the cosine similarity of output in the previous layer.

\subsection{Adaptations made for the evaluation}
The GraphMatcher's framework has been developed in Python with PyTorch and Ontospy, and is packed by SEALS using MELT~\cite{Sven_melt_2010}.

\subsection{Parameter settings}
The model's parameters~\footnote{These parameters do not give the optimum model. The best model has different parameters, but we mistakenly submitted the model using these parameters to the conference track challenge.} are 0.01 of learning rate, 5 epochs, 0.001 of weight decay and 16 of batch size. The threshold is computed from false positive alignments in the validation data as how the VeeAlign~\cite{VeeAlign} system proposes~\footnote{The project uses VeeAlign's approach directly to compute the threshold with the permission of the first author.}.
\section{Results}
The conference track consists of sixteen ontologies, but only seven ontologies have (twenty-one) reference alignment cases. These reference alignments have been utilized as ground truths to use true positive alignments. Besides, negative alignment cases have been computed by oversampling from all the possible class and property alignments. Therefore, once we apply a supervised machine learning approach, we can use only these seven ontologies from this dataset.

\begin{table}[!h]
    \centering
\caption{The table shows GraphMatcher results in the conference track.}
\begin{tabular}{l||l|l|l|l|l}
\toprule
\textbf{Evaluation} & \textbf{Precision} & \textbf{F.5-measure} & \textbf{F1-measure} & \textbf{F2-measure} & \textbf{Recall} \\
\midrule
\textbf{ra1-M1}& 0.82 & 0.77& 0.71 & 0.65 & 0.62 \\
\textbf{ra1-M2}& 0.65 & 0.51 & 0.39 & 0.32 & 0.28 \\
\textbf{ra1-M3}& 0.8 & 0.74 & 0.67 & 0.6 & 0.57 \\
\textbf{ra2-M1}& 0.78  & 0.73 & 0.66 & 0.6 & 0.57 \\
\textbf{ra2-M2}& 0.65 & 0.5 & 0.39 & 0.32 & 0.28\\
\textbf{ra2-M3}& 0.76 & 0.7 &  0.62 & 0.56 & 0.53 \\
\textbf{rar2-M1}& 0.77 & 0.73 & 0.67 & 0.62 & 0.59 \\
\textbf{rar2-M2}& 0.65 & 0.52 & 0.4 & 0.33 & 0.29\\
\textbf{rar2-M3}& 0.75 & 0.7 & 0.63& 0.58 & 0.55\\
\toprule
\end{tabular}
    \label{tab:conference_result}
\end{table}

The Table~\ref{tab:conference_result} shows the GraphMatcher results in the conference track. The performance of the GraphMatcher is better in M1 variants than in M2 in terms of F1-measure. Therefore, its F1-measure has been decreased in M3 variants. As a result, the model has a weakness in the M2 variants, namely property alignments. 
\section{General Comments}
\subsection{Comments on the results}
The GraphMatcher is the new ontology matching system participating in OAEI 2022 and is evaluated in the conference track. The GraphMatcher demonstrates remarkable performance in the M1 and M3 evaluation variants in terms of F1-measure, even though it does not have high performance in the M2 evaluation variant. However, all other matchers do not show remarkable results in this M2 evaluation variant. 

On the other hand, it is also evaluated in the uncertain reference alignments in OAEI 2022 conference track. It has the highest F1-measure (72\% in both of them) in the discrete and continuous metrics~\footnote{The evaluation based on the uncertain reference alignments:~\url{http://oaei.ontologymatching.org/2022/results/conference/index.html}} among all other matchers evaluated in this track. This means that the GraphMatcher's confidence is higher than the other matchers evaluated in the OAEI 2022 conference track.

\begin{table}[!h]
    \caption{The table shows the matchers' performances on the rar2-M3 reference alignments.}
    \begin{tabular}{l||l|l|l|l|l}
    \toprule
    \multicolumn{1}{l|}{\textbf{Matcher}} & \multicolumn{1}{l|}{\textbf{Precision}} & \multicolumn{1}{l|}{\textbf{F.5-measure}} & \multicolumn{1}{l|}{\textbf{F1-measure}} & \multicolumn{1}{l|}{\textbf{F2-measure}} & \multicolumn{1}{c}{\textbf{Recall}} \\ 
    \midrule
        LogMap & 0.76 & 0.71 & 0.64 & 0.59 & 0.56 \\
        \textbf{GraphMatcher} & 0.75 & 0.7 & 0.63& 0.58 & 0.55 \\
        SEBMatcher & 0.79 & 0.7 & 0.6 & 0.52 & 0.48 \\
        ATMatcher & 0.69 & 0.64 & 0.59 & 0.54 & 0.51 \\
        ALIN & 0.82 & 0.7 & 0.57 & 0.48 & 0.44 \\
        LogMapLt & 0.68 & 0.62 & 0.56 & 0.5 & 0.47 \\
        LSMatch & 0.83 & 0.69 & 0.55 & 0.46 & 0.41 \\
        AMD & 0.82 & 0.68 & 0.55 & 0.46 & 0.41 \\
        KGMatcher+ & 0.83 & 0.67 & 0.52 & 0.43 & 0.38 \\
        ALIOn & 0.66 & 0.44 & 0.3 & 0.22 & 0.19 \\
        Matcha & 0.37 & 0.2 &0.12 & 0.08 & 0.07 \\
    \bottomrule
    \end{tabular}
\end{table}
\subsection{Improvements}
The GraphMatcher should be improved to match the properties, since it does not perform well in the M2 evaluation variants. Its current version does not apply the graph attention to align the properties because of the lack of properties' neighbours, especially datatype properties. These object type and data type properties might not have enough neighbouring terms to construct contextual information in the ontology. In this case, the property's context might be improved with the external information in its next version, and the graph attention can also be applied to align the properties. In addition to these improvements, we will train the model with its optimum parameter settings in its further version.

\section{Conclusion}
In this study, we have introduced the new ontology-matching system called GraphMatcher. The GraphMatcher adapted the graph attention to the homogeneous subgraphs of the centre class' neighbours to obtain the contextual information about the centre class.
The graph attention has computed the higher-level representation of each class and its surrounding classes and properties. The results demonstrate promising performances in M1 and M3 evaluation variants. The future work of this study will be to increase its performance in M2 evaluation variants.


\bibliography{references}
\end{document}